\documentclass{article}

    \PassOptionsToPackage{numbers, compress}{natbib}

\usepackage[preprint]{neurips_2022}

\usepackage{amsmath}
\usepackage[utf8]{inputenc} 
\usepackage[T1]{fontenc}    
\usepackage{hyperref}       
\usepackage{url}            
\usepackage{booktabs}       
\usepackage{amsfonts}       
\usepackage{nicefrac}       
\usepackage{microtype}      
\usepackage{xcolor}         
\usepackage{graphicx}
\usepackage{multirow}

\title{Efficient Few-shot Identity Preserving Attribute Editing for 3D-aware Deep Generative Models}

\author{%
  Vishal Vinod\\
  Department of Computer Science\\
  UC San Diego\
  La Jolla, CA 92093 \\
  \texttt{vvinod@ucsd.edu} \\
}

\begin{document}

\maketitle

\begin{abstract}
    Identity preserving editing of faces is a generative task that enables modifying the illumination, adding/removing eyeglasses, face aging, editing hairstyles, modifying expression etc., while preserving the identity of the face. Recent progress in 2D generative models have enabled photorealistic editing of faces using simple techniques leveraging the compositionality in GANs \cite{parihar2022everything, chai2021using}. However, identity preserving editing for 3D faces with a given set of attributes is a challenging task as the generative model must reason about view consistency from multiple poses and render a realistic 3D face. Further, 3D portrait editing requires large-scale attribute labelled datasets and presents a trade-off between editability in low-resolution and inflexibility to editing in high resolution. In this work, we aim to alleviate some of the constraints in editing 3D faces by identifying latent space directions that correspond to photorealistic edits. Recent 3D-aware methods utilizing the inherent semantic hierarchy for editing require semantic masks and training with expensive volumetric rendering at high resolutions with up-sampling that breaks multi-view consistency. While there has been research on conditional semantic space editing and efficient 3D-aware GAN inversion techniques, there has been minimal attention toward highly controlled identity preserving editing in 3D. To address this, we present a method that builds on recent advancements in 3D-aware deep generative models and 2D portrait editing techniques to perform efficient few-shot identity preserving attribute editing for 3D-aware generative models. We aim to show from experimental results that using just ten or fewer labelled images of an attribute is sufficient to estimate edit directions in the latent space that correspond to 3D-aware attribute editing. In this work, we leverage an existing face dataset with masks to obtain the synthetic images for few attribute examples required for estimating the edit directions. Further, to demonstrate the linearity of edits, we investigate one-shot stylization by performing sequential editing and use the (2D) Attribute Style Manipulation (ASM) \cite{parihar2022everything} technique to investigate a continuous style manifold for 3D consistent identity preserving face aging. Code and results are available at: \href{https://vishal-vinod.github.io/gmpi-edit/}{https://vishal-vinod.github.io/gmpi-edit/}
    
    \medskip
      


    
    

    

\end{abstract}


\section{Introduction}
\label{sec:intro}
    \paragraph{Problem Definition.} We aim to find disentangled latent space edit directions in 3D-aware deep generative models \cite{chan2022efficient, zhao2022generative, sun2022ide, sun2022fenerf} that enable highly controlled identity preserving attribute editing while also preserving multi-view consistency. Multi-view consistency represents 3D fidelity in face image generation - the generated images must be photorealistic when rendered from various camera poses despite being trained on single-view image collections. As described in Fig.(\ref{fig:model}), StyleGANv2 \cite{karras2019style} is used as the 2D image prior with the alpha-map generator to generate a multi-plane image followed by MPI (Multi-Plane Image) rendering conditioned on the camera to obtain the 3D and multi-view consistent face portrait. Here, the inversion network and latent edit computation provide the latent offsets that correspond to the attribute of interest change in the rendered image. StyleGANv2 consists of a mapping network that maps a randomly sampled z-latent to a 512-dimensional $\mathcal{W}$-latent space. The $\mathcal{W}$ latent is replicated for each synthesis block of the StyleGANv2 generator network to generate a 2D image. Note that for training at 256x256 resolution, there are t=14 synthesis blocks and for training at 1024x1024 resolution there are t=18 synthesis blocks. Hence we replicate the $\mathcal{W}$-latent vector t number of times based on the training resolution of our model. This vector of shape $(t, 512)$ constitutes the $\mathcal{W}+$ latent space of StyleGANv2. We use the computed edit direction for the attribute by simply adding it to the $\mathcal{W}$ latent vector of StyleGANv2. The StyleGANv2 network replicates the $\mathcal{W}$ latent vector into the $\mathcal{W}+$ latent space and propagates the attribute information through the generator at all resolution scales. We draw on insights from extensive 2D portrait editing literature and investigates their applicability in 3D-aware GANs. Several prior works including recent methods FeNeRF \cite{sun2022fenerf}, IDE-3D \cite{sun2022ide} utilize the semantic hierarchy of 3D-aware deep generative models to perform edits. Ours is the first work that tests the hypothesis of directly estimating latent space edit directions eliciting 3D consistent and identity preserving edits. The focus on efficiency and few-shot attribute editing enables portrait editing using few labelled examples of the attribute from a collection of synthetic data. Further, the investigation of zero-shot stylization \cite{chong2021jojogan} from sequential edits and continuous style manifold for face aging are to consolidate the effectiveness of latent space edits in 3D-aware deep generative models.

\paragraph{Problem Significance.}
Identity preserving editing is a challenge in 3D-aware generative models because the latent space is pose conditioned \cite{chan2022efficient} and highly entangled. Semantic 2D editing methods require large-scale attribute labelled datasets and attribute classifiers and include a trade-off between editability and resolution. Further, a recent method IDE-3D \cite{sun2022ide} while enabling semantic 3D consistent real-time editing, requires training with expensive volumetric rendering at high resolutions with up-sampling that breaks multi-view consistency resulting in low fidelity geometry. By enabling few-shot identity preserving 3D consistent editing by estimating latent space edit directions, we alleviate (1) the requirement of large scale labelled datasets, (2) expensive computation from methods employing volumetric rendering during training \cite{sun2022ide, sun2022fenerf}, and (3) the editability-resolution trade-off. To address this, the significance of our contribution is as follows:
\begin{enumerate}
    \item A 3D consistent few-shot identity preserving attribute editing formulation utilizing only up to ten synthetic image pairs for an attribute. Identifying disentangled edit directions enabling 3D consistent identity preserving sequential edits.
    \item Investigating sequential edits to demonstrate the disentangled identity preserving latent space edit directions.
    \item Perform Pivotal tuning inversion (PTI) \cite{roich2022pivotal} to invert out-of-distribution images into the GAN latent space to generate 3D consistent orbits. This enables the rendering of 3D faces without camera poses. (In domain inversion follows \cite{zhu2020domain}).
\end{enumerate}

\begin{figure}
    \centering
    \includegraphics[width=0.99\linewidth]{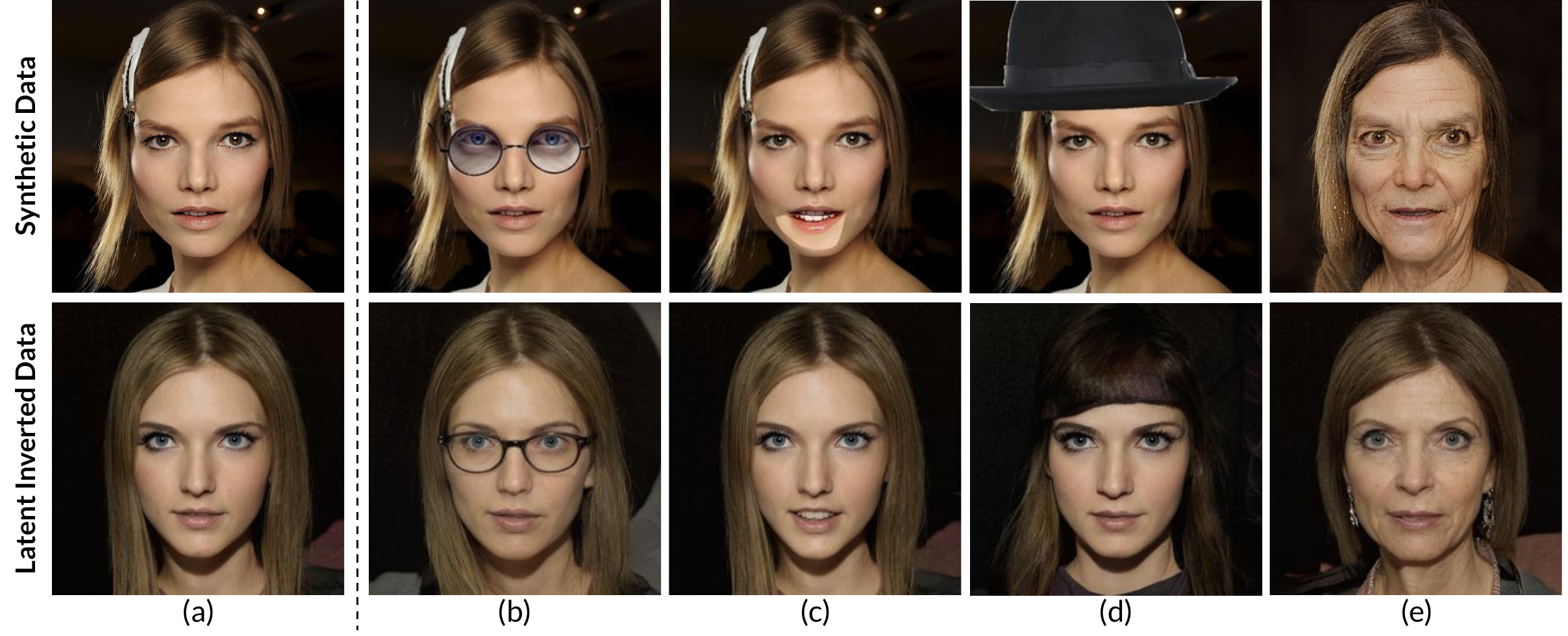}
    \caption{Experimental results for out-of-distribution image inversion to demonstrate the robustness of the style transfer across datasets using the synthetic image pairs. Here, the first row represents the synthetic image data and the second row shows the latent inversion results. Column (a) represents the input images for identity 3 in the CelebA-HQ dataset. Columns (b), (c), (d) and (e) represents the edits corresponding to eyeglasses, expression, hat and face aging respectively. Notice how the positive images (row 1) appear unrealistic, however, the inversion is in a continuous space demonstrating the GAN's ability to be invariant to inaccuracies when mapping to the latent space demonstrating the underlying hypothesis of the approach making it suitable for multi-view consistency.}
    \label{fig:inv_2}
    \vspace{-2mm}
\end{figure}

\paragraph{Technical Challenge}
There are several deficiencies in current methods for editing 3D-aware GANs. The requirement of large-paired datasets for attribute classification in several 2D editing methods \cite{alaluf2021only, zhu2021barbershop, khodadadeh2022latent, gao2021high, liang2021ssflow} is difficult as masked segmentation of fine grained face attributes such as hairstyle and lighting conditions \cite{zhou2019deep} are difficult to label. While recent methods such as DatasetGAN \cite{zhang2021datasetgan} aim to minimize human effort by introducing a few-shot labelling pipeline, the task of labelling subjective and identity preserving attributes such as age is not possible. To alleviate this requirement, we reduce the dependence on paired examples and large scale datasets by utilizing a few-shot synthetic data approach based on FLAME \cite{parihar2022everything}. Using this formulation, we only require up to ten images with the attribute under consideration to estimate disentangled edit directions in 3D-aware GANs. We only require images with the attribute of interest and a mask for the attribute of interest to create a set of synthetic pairs of positives (images with the cut-and-pasted attribute) and negatives (images without the attribute). These sets of synthetic pairs of images are inverted into the GAN latent space to identify the $\mathcal{W}+$ latent space edit directions (refer Sec.(\ref{subsec:dataset})). In Fig.(\ref{fig:inv_2}), the first row depicts the synthetic data we create using the negatives along with the positives with the cut-and-pasted attribute from the attribute images. Note that for the old age edit, we use an off-the-shelf aging network \cite{alaluf2021only} to obtain the aged version of the identity as there is no applicable masked style for age. However, for other styles such as the eyeglasses, expression and hat, we conform to our cut-and-paste approach. 

Recent methods for editing 2D GANs have used the semantic information modeled \cite{park2019semantic, zhu2020sean} to elicit effective, realistic and real-time editing results \cite{ling2021editgan, zhang2021datasetgan}. Recent 3D aware GAN editing methods such as SofGAN \cite{chen2022sofgan}, FENeRF \cite{sun2022fenerf}, IDE-3D \cite{sun2022ide} and explicitly controlled editing \cite{tang2022explicitly} learn a semantic editing space that enable editing the 3D representation using the semantic hierarchy inherently modeled by the 3D-aware deep generative model. SofGAN \cite{chen2022sofgan} disentangles the latent space into geometry and texture sub-spaces to edit 3D representations encoded as semantic occupancy fields (semantic 3D volumes). FENeRF \cite{sun2022fenerf} extends the $\pi$-GAN \cite{chan2021pi} method to learn disentangled semantics and aligned textures leveraging the 3D representation to enable 3D editing. IDE-3D \cite{sun2022ide} learns disentangled representations using semantic and texture tri-planes while explicitly controlled editing \cite{tang2022explicitly} uses tri-planes and a 3DMM prior to edit the semantic space eliciting edits in the 3D representation using volume blending. However, as stated, these methods depend on strong priors (3DMM) and need to be trained with rendered semantic masks and require expensive volumetric rendering which includes large overheads especially at high resolutions.

Challenges in implementation and engineering in current 3D-aware generative models particularly deal with the expensive volumetric rendering during training. ShadeGAN \cite{pan2021shading} proposes a novel surface tracking method that enables a $24\%$ improvement during training by learning a lightweight CNN network that predicts the depth (learnt using a depth-mimic loss) in order to reduce the number of points used for rendering. Several engineering challenges exists especially when the handedness of the camera or other camera intrinsics are unknown and non-transferable across datasets. Evaluation of editing in GANs typically involve comparison across metrics such as FID (Frechet Inception Distance), KID (Kernel Inception Distance). To evaluate the identity preserving capability, ID (multi-view identity consistency) has been used. However, in this work, we work on several editing attributes including aging wherein an analysis including a qualitative study where participants rank different edits across the baselines and proposed method may provide better insights on photorealism.




\section{Related Work}
\label{sec:related}
    \textbf{3D-aware generative models.} Learning 3D representations from multi-view images and camera poses have been extensively studied following the explosion of Neural Radiance Fields (NeRFs) \cite{mildenhall2021nerf} and Neural Fields \cite{srinivasan2021nerv, zhang2020nerf++, barron2021mip, zhang2021nerfactor, gu2021stylenerf}. However, these methods learn a radiance field for a single scene, require several views of the same scene and make use of expensive volumetric rendering at every training step making the task of generation at high resolution prohibitively expensive. While RegNeRF \cite{niemeyer2022regnerf} reduces the need for several views of the same scene to only a handful of images, the results have artifacts and still require volumetric rendering at every train step. Recent works aim to reduce the need for several views of the same scene by utilizing the canonical space afforded by faces to learn 3D representations from a collection of single-view images captured from arbitrary viewing directions including $\pi$-GAN \cite{chan2021pi}, EG3D \cite{chan2022efficient, lin20223d}, EpiGRAF \cite{skorokhodov2022epigraf}, LoLNeRF \cite{rebain2022lolnerf}, TEGLO-NeRF \cite{vinod2024teglo, vinod2023neural} and GMPI \cite{zhao2022generative}. While these methods can generate 3D consistent faces (and ShapeNet objects \cite{chang2015shapenet}) and interpolate between identities, they do not allow controllable editing capabilities. Two recent works: FENeRF \cite{sun2022fenerf} and IDE-3D \cite{sun2022ide}, exploit the semantic hierarchy in 3D-aware GANs to allow semantic space editing (IDE-3D enables real-time semantic editing), however, they require training with semantic masks and expensive volumetric rendering which is expensive at high resolutions. Explicitly Controllable 3D editing \cite{tang2022explicitly} utilizes the semantic space and a 3DMM \cite{blanz1999morphable} prior to enable editing but still require volumetric rendering and several moving parts to enable editing. In this work, we propose the first latent space editing exploration in 3D-aware GAN method GMPI \cite{zhao2022generative} that exploits the compositionality property \cite{chai2021using} to find latent space edit directions using only the pre-trained 3D-aware GAN.

\textbf{2D GAN editing.} Stylization in 2D involves several techniques such as zer-shot stylization \cite{chong2021jojogan}, contrastive style transfer \cite{park2020contrastive}, two-stream AdaIn-based stylization \cite{jiang2020tsit} etc. Conditional editing in 2D GANs require large-scale paired datasets inclusive of all attributes for attribute classification. Editing the $\mathcal{W}/\mathcal{W}+$ latent space (described in Sec.(\ref{sec:intro})) has been explored in 2D GANs \cite{parihar2022everything, abdal2021styleflow, shen2020interfacegan, harkonen2020ganspace, yuksel2021latentclr}. InterFaceGAN \cite{shen2020interfacegan} finds edit direction using SVM in the latent space, GANSpace \cite{harkonen2020ganspace} finds edit directions using PCA on latent codes along with manual filtering, StyleFlow \cite{abdal2021styleflow} uses continuous normalized flow for latent space transformations, FLAME \cite{parihar2022everything} finds edit directions using PCA on the latent difference of samples. 2D GANs exploiting the semantic hierarchy such as EditGAN \cite{ling2021editgan} use the segmentation mask from a GAN jointly trained to generate the image and mask in order to enable GAN editing. While several methods perform 2D GAN editing by identifying latent space modifications \cite{alaluf2021only, zhu2021barbershop, khodadadeh2022latent, gao2021high, liang2021ssflow}, they require attribute classifiers and identify entangled edit directions that do not allow controlled and identity preserving edits. Moreover, these methods only explore the 2D GAN editing capabilities whereas the editing landscape in 3D is challenging due to the requirement of multi-view consistency with large pose variations.

\section{Proposed Solution}
\label{sec:method}
    \begin{figure}
    \centering
    \includegraphics[width=0.99\linewidth]{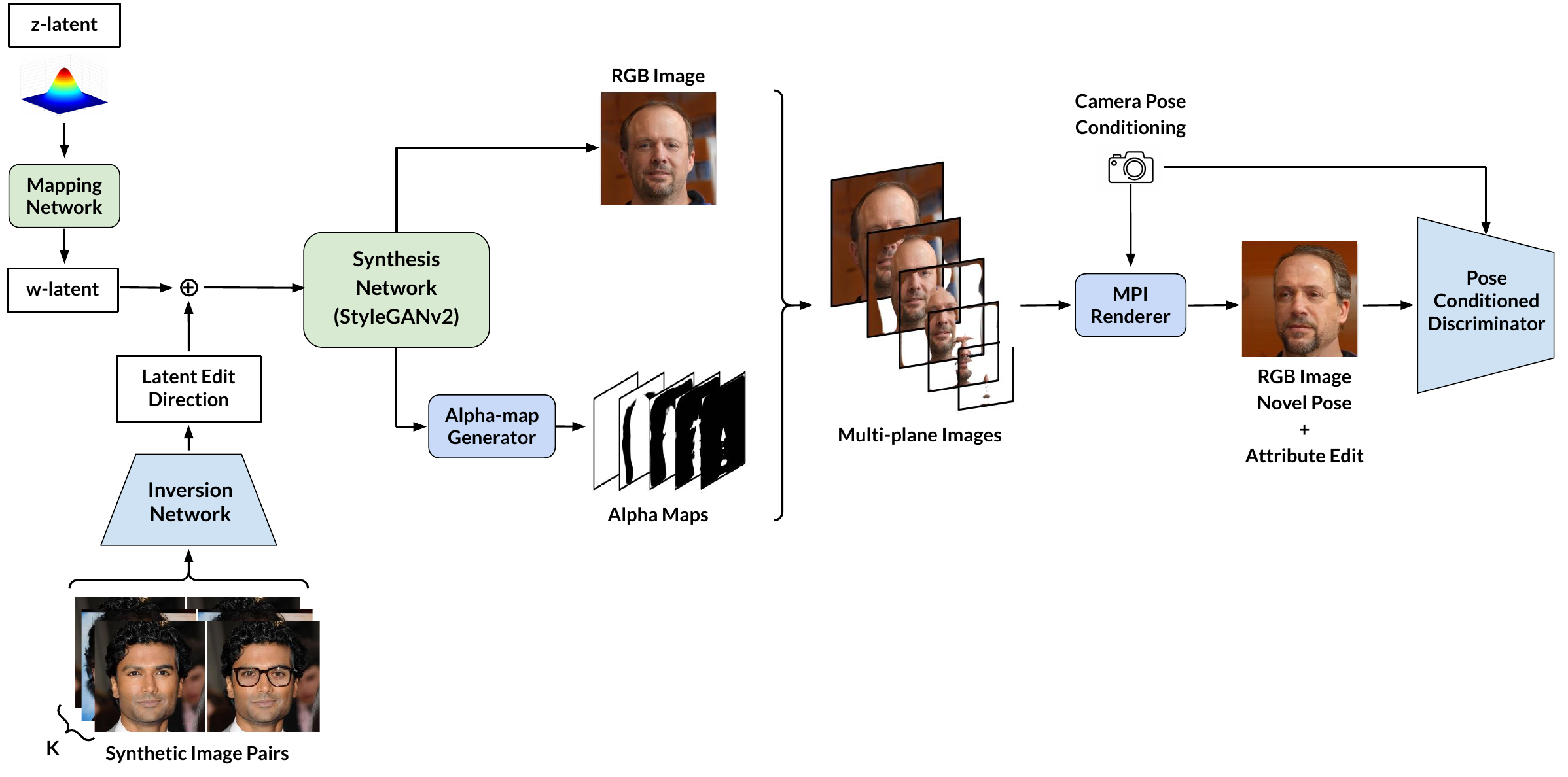}
    \caption{Proposed model architecture. On the left half we have the components of the StyleGANv2 model components (Mapping network, StyleGANv2 generator). In addition we have the Inversion network (depicted in blue at the bottom left). There are up to k pairs of synthetic image pairs created from the CelebA-HQ dataset using the CelebA-HQ Mask dataset's corresponding image masks. The latent difference direction of the inverted image is added to the w-latent of the StyleGANv2 network and passed to the Alpha Map generator (depicted in blue on the left half). Further, the combined results are passed to the MPI (Multi-Plane Image) renderer conditioned on an arbitrary camera pose and finally the rendered output is passed to the pose conditioned discriminator. In the above example we expect the rendered face to also posses the attribute of interest i.e the eyeglasses from the synthetic image pairs (bottom left k image pairs).}
    \label{fig:model}
    \vspace{-1mm}
\end{figure}

\paragraph{Idea Summary.}
In this work, we present a method to perform efficient few-shot identity preserving attribute editing for 3D-aware generative models. We propose to show from experimental results using just ten or fewer labelled images of an attribute, we can estimate disentangled edit directions in the latent space that correspond to 3D-aware attribute editing while also preserving identity. We aim to demonstrate the disentangled nature of the edits despite the challenges in multi-view consistency and large pose variability by showing sequential edits and one-shot stylization of 3D representations. Further, we experiment with continuous style space edits such as for face aging in order to investigate the strength of the proposed method. Lastly, we also investigate inversion and pivotal tuning inversion to invert out-of-distribution image samples into the GAN latent space. 


\begin{figure}
    \centering
    \includegraphics[width=0.99\linewidth]{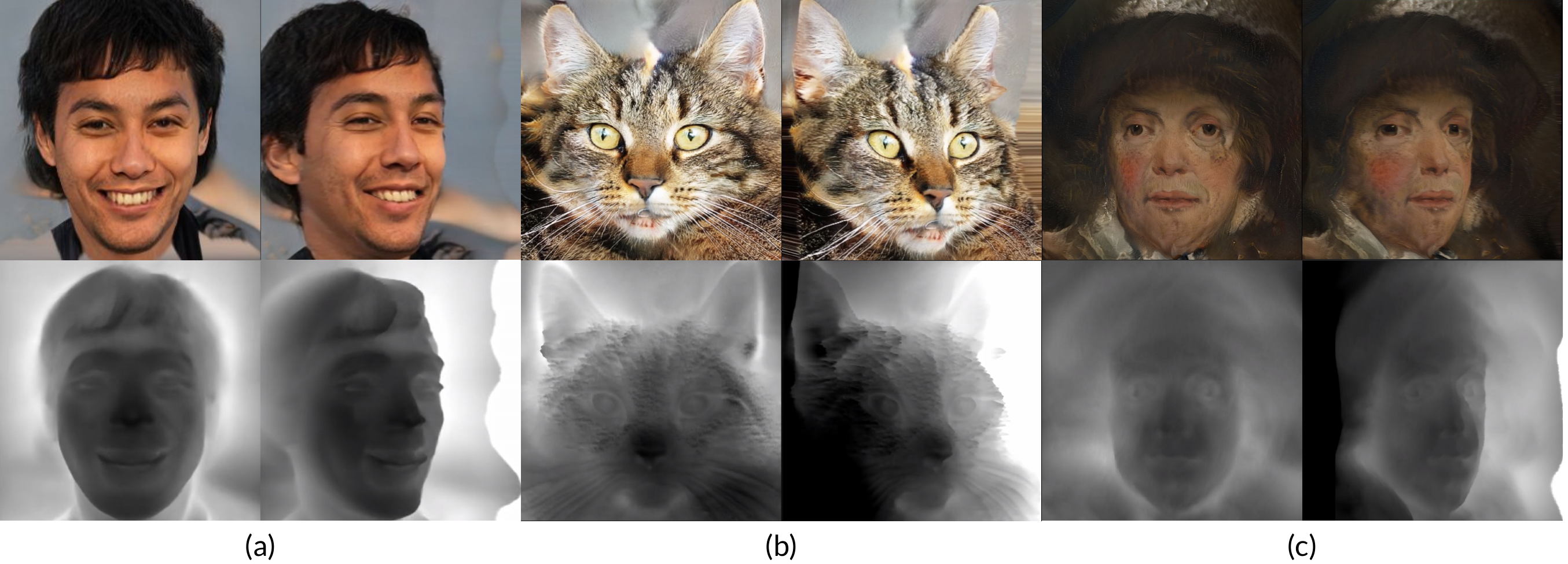}
    \caption{Qualitative multi-view consistent results with GMPI trained on (a) FFHQ dataset, (b) AFHQv2 Cats dataset, (c) MetFaces dataset \cite{karras2020training}. (Row 1) depicts the RGB images demonstrating camera conditioned MPI rendering, (Row 2) depicts the corresponding depth maps for the RGB images.}
    \label{fig:ffhq_prelim}
\end{figure}

\paragraph{Problem setting and description.} 
As denoted in Fig.(\ref{fig:model}), the proposed method follows from the GMPI model along with a inversion network that takes as input $k$ synthetic image pairs to determine the attribute edit direction. As described in Sec.(\ref{sec:intro}), StyleGANv2 consists of a mapping network that maps a randomly sampled 512-dimensional z-latent to a 512-dimensional $\mathcal{W}$-latent space (Fig.(\ref{fig:model})). The $\mathcal{W}$ latent is replicated for each synthesis block of the StyleGANv2 generator network to generate a 2D image. Note that for training at 256x256 resolution, there are t=14 synthesis blocks and for training at 1024x1024 resolution there are t=18 synthesis blocks. Hence, we replicate the $\mathcal{W}$ -latent vector t number of times based on the training resolution of our model. This vector of shape $ (t, 512) $ constitutes the $\mathcal{W}+$ latent space of StyleGANv2. We use the computed edit direction for the attribute by simply adding it to the $\mathcal{W}$ latent vector of StyleGANv2. The StyleGANv2 network replicates the $\mathcal{W}$ latent vector into the $\mathcal{W}+$ latent space and propagates the attribute information through the generator at all resolution scales. Further, the alpha-map generator and camera-conditioned MPI renderer are used to obtain the multi-plane image output in a novel pose. The hypothesis is based on the compositionality property of GANs from the findings of \cite{chai2021using} where the authors show that inverting a pre-trained generator is analogous to providing a modified latent (in a smooth latent space) to the strong prior (generator) and even if the input to the inversion network is unrealistic, the combination of the inversion network and the generator map the input into a manifold that renders a realistic output. To train the inversion encoder, we keep the StyleGANv2 network, alpha-map generator and the MPI renderer frozen and reconstruct the output pose-conditioned image and the corresponding $\mathcal{W}+$ latent (described above) to render the image. The inversion network is trained with the latent reconstruction loss, input image reconstruction loss and the LPIPS loss \cite{zhang2018unreasonable} for photo-realistic inversion. In the following equation, $\mathcal{L}_{\text{LPIPS}}$ is the Learned Perceptual Image Patch Similarity loss, $\mathcal{I}$ is the inversion network, $\mathcal{S}_k$ is the set of $k$ synthetic image pairs, $\mathcal{G}$ is the GMPI generator network comprising of the StyleGANv2 generator, alpha-map generator and the MPI renderer. The latent reconstruction loss is the $\mathcal{L}_2$ loss between the modified $\mathcal{W}+$ latent and the $\mathcal{W}+$ latent from the mapping network obtained by sampling a $z$-latent from the Gaussian distribution. In Eq(2), x is the ground truth input image that is being reconstructed using inversion. Eq(1) represents the LPIPS loss where H is the image height, W is the image width, $\widehat{y}_{hw}^l$ and $\widehat{y}_{0hw}^l$ are the outputs from l feature layers of the feature extractor, $v_l$ is the activation scale vector. The LPIPS loss is given below:  

\begin{equation}
   \mathcal{L}_{\text{LPIPS}} = \sum_l \frac{1}{H \times W} \times \sum_{h, w} || v_l \times (\widehat{y}_{hw}^l - \widehat{y}_{0hw}^l) ||_2^2 
\end{equation}

The total loss to train the inversion network is given below (note that the GMPI model including the MPI renderer is frozen and not trained as we use it as a strong prior with completion and composition properties).

\begin{equation}
    \mathcal{L}_{\text{inv}} = \lambda_{\text{LPIPS}} \times \mathcal{L}_{\text{LPIPS}} + \lambda_{\text{recons}} \times || \; x - \mathcal{G}(\text{I}(\mathcal{S}_k)) ||_2 + \lambda_{\text{latent}} \times || w - \mathcal{W+} \; ||_2
\end{equation}



To compute the edit directions for an attribute, we first create $k$-synthetic image pairs (upto 10 image pairs) from the CelebA dataset using the attribute masks from the CelebAMask-HQ dataset. Thus for each of the attributes in the CelebAMask-HQ dataset, we obtain a synthetic dataset pair of $k$ samples. Here, we denote the image itself as the negative and the image with the attribute overlay as the positive. We then pass these images through the inversion network $\mathcal{I}$ and obtain the latent directions. We then compute the difference between the positive latent directions and negative latent directions and compute the SVD (Singular Value Decomposition) of the difference in order to obtain the consolidated edit direction for that attribute. This is then propagated through the GMPI network by adding the latent direction to the $\mathcal{W}+$ latent and the corresponding edited face is rendered using the MPI (Multi-Plane Image)renderer. 

\textbf{Implementation.} We use the PyTorch deep learning framework for our experiments. The GMPI model is trained at $1024 \times 1024$ resolution using a (frozen) pre-trained StyleGANv2 for the MPI rendering on the FFHQ dataset. For qualitative results for MetFaces and AFHQv2, we use $256 \times 256$ resolution trained model from the author's implementation \href{https://github.com/apple/ml-gmpi}{[Link]}. As we use a $1024$ resolution model for the experiments with FFHQ, our $\mathcal{W}+$ latent space has t=18 blocks. Further, for the inversion network, the Adam \cite{kingma2014adam} optimizer is used for training along with weighted losses: LPIPS loss, latent reconstruction loss and the image reconstruction loss (refer Eq.2). For experiments involving pivotal tuning inversion, the identity loss from \cite{richardson2021encoding} and the network is trained at $256 \times 256$ resolution referenced from the author's codebase \href{https://github.com/chail/latent-composition}{[Link]}. Experiments with MPI rendering at $1024 \times 1024$ was performed using a NVIDIA Tesla v100 (32 GB GPU) and the latent direction estimation and related experiments for pivotal tuning inversion were performed using a NVIDIA Tesla T4 (16 GB GPU). 

\begin{figure}
    \centering
    \includegraphics[width=0.8\linewidth]{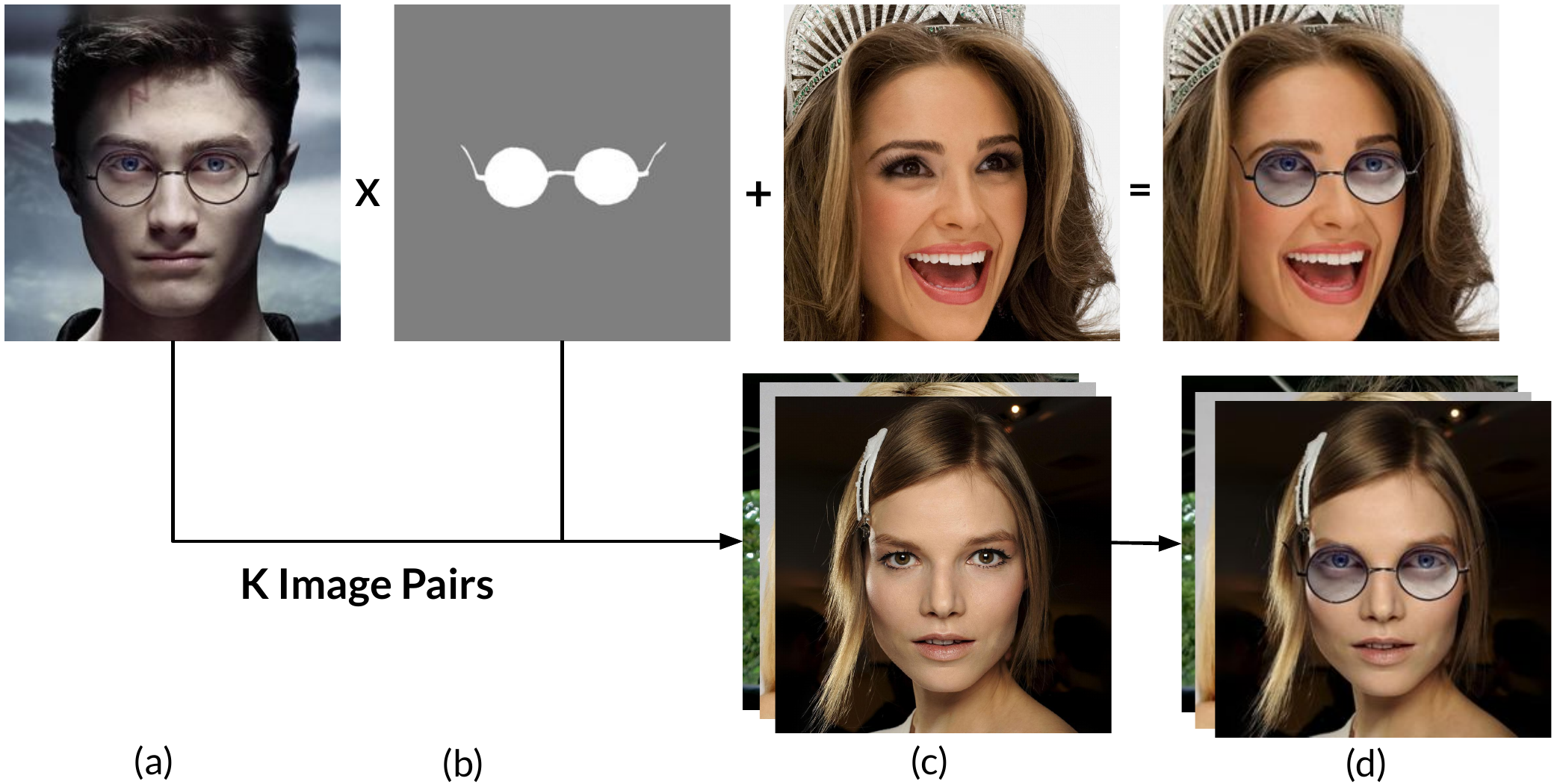}
    \caption{Pictorial demonstration of synthetic image pair creation for inversion. Here (a) represents the ground truth image with the attribute of interest - Eg: eyeglasses, (b) The mask from CelebA-HQ Mask is used to extract the eyeglasses. (c) Negative image i.e any image without the attribute of interest, and (d) Positive image: resulting image from applying cut-and-paste fro the attribute image.}
    \label{fig:synth}
    \vspace{-2mm}
\end{figure}




\section{Experiments and Results}
\label{sec:results}
\begin{figure}
    \centering
    \includegraphics[width=0.99\linewidth]{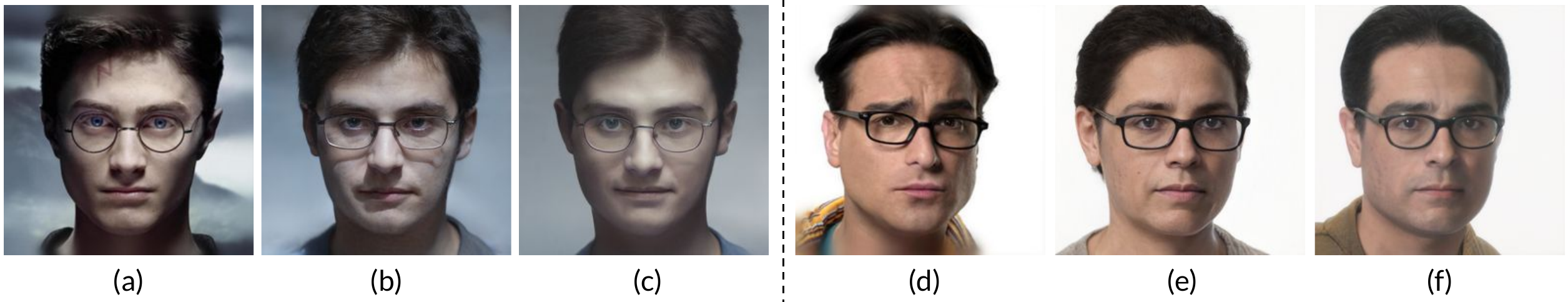}
    \caption{Experimental results for our investigation of inverting out-of-distribution (OOD) images into the method's latent space. In this experiment we use a model trained on the FFHQ dataset and test inversion with images from the CelebA-HQ dataset. In the figure: (a) Ground truth image from the CelebA-HQ dataset, (b) Direct latent inversion with no fine-tuning (c) Pivotal tuning inversion (fine-tuning) in the latent space. Similarly, (d), (e) and (f) shows an image from the CelebA-HQ dataset, direct latent inversion, and pivotal tuning inversion of the image respectively. }
    \label{fig:inv}
\end{figure}

    \paragraph{Datasets and Tools.}
\label{subsec:dataset}
To create the synthetic dataset of positive and negative pairs, we chose an image $\mathcal{S}$ as the source identity. Next, we identify up to images in the CelebAMask-HQ \cite{CelebAMask-HQ} dataset that correspond to a style attribute denoted as $\mathcal{M}_1, \mathcal{M}_2, ..., \mathcal{M}_{10}$. We use the corresponding attribute's mask for facial attributes (such as skin, nose, eyes, eyebrows, ears, mouth, lip, hair, hat, eyeglass, earring, necklace, neck, and cloth) to obtain the attribute masked out as output. This is then pasted onto $\mathcal{S}$ so as to obtain a positive image $\mathcal{P}$. The source identity $\mathcal{S}$ without the super-imposed style attribute is the negative. We create up to ten such positive-negative pairs for an attribute. Note that the super-imposed positive image $\mathcal{P}$ will likely look unrealistic, however, the compositional property of an encoder with a strong prior GAN in GAN inversion enables the generation of a realistic representation despite an unrealistic input as it inverts the input to a latent that generates realistic outputs. Note that for an attribute such as age which cannot posses a mask, we use an off-the-shelf aging generator \cite{alaluf2021only} to generate the aged version (shown in Fig.(\ref{fig:inv})). The CelebA-HQ \cite{liu2015faceattributes, karras2017progressive}: A dataset of 30,000 images of faces at $1024\times1024$ resolution and CelebAMask-HQ \cite{CelebAMask-HQ}: A dataset of 30,000 images of faces at $512 \times 512$ resolution including 19 categorical semantic labels including all facial components and accessories such as skin, nose, eyes, eyebrows, ears, mouth, lip, hair, hat, eyeglass, earring, necklace, neck, and cloth are used to create the synthetic dataset. We aim to use the following datasets for the experiments in this work: \begin{enumerate}
    \item AFHQv2 Cats \cite{karras2021alias, choi2020stargan}: a real-world dataset of cat faces comprising $6,000$ faces at $512 \times 512$ resolution RGB images. Off-the-shelf pose estimators \cite{deng2019accurate, cathipster} used to obtain camera intrinsics.
    \item MetFaces \cite{karras2020training}: A dataset of 1,336 art portrait face images at $1024\times1024$ resolution downloaded from the Metropolitan Museum of Art Collection API, that is aligned and cropped.
    \item FFHQ \cite{karras2019style}: a real-world single-view human face dataset comprising 70,000 faces and camera poses at $1024 \times 1024$ resolution RGB images. Off-the-shelf pose estimators \cite{deng2019accurate, cathipster} are used to obtain camera approximate extrinsics.
\end{enumerate}

\begin{figure}
    \centering
    \includegraphics[width=0.85\linewidth]{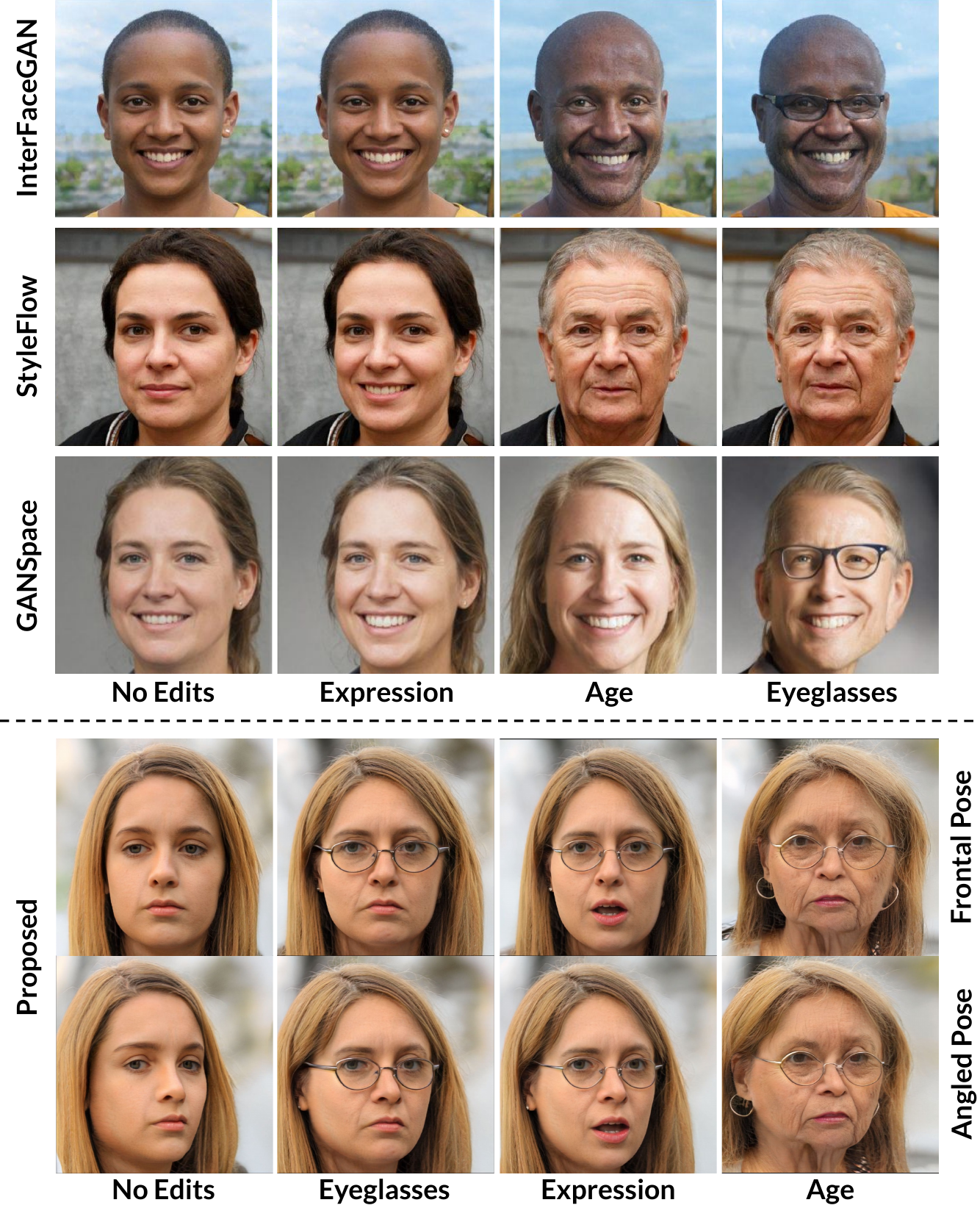}
    \caption{Qualitative comparison of sequential edits with previous methods along with multi-view pose consistency of our method. Each row corresponds to the method's result for the attribute edit indicated and a sequential edit indicates the addition of edit latent vectors (described in Sec.(\ref{sec:results})). Baseline results obtained from \cite{parihar2022everything}: For InterFaceGAN we observe that the aging edit changes the identity of the person. For StyleFlow we observe that the aging edit completely changes the identity and is unable to add in 3D geometry altering edits such as eyeglasses. Similarly, for GANSpace, we have the aging and eyeglass edits modifying the identity of the person. However, in our method, we demonstrate identity-preserving 3D consistent editing at 1024x1024 resolution with disentangled edits. Notice the geometry of the eyeglasses is consistent across multiple camera views. The mouth expression edit (col. 3) shows 3D consistent photorealism and age (age=70) edit shows 3D consistent wrinkles. We show frontal and angled pose for our method to demonstrate multi-view consistency.}
    \label{fig:sequential}
\end{figure}

\paragraph{Baselines.}
We compare with previous state-of-art latent editing methods in 2D GANs and aim to extrapolate their performance for 3D-aware GANs. InterfaceGAN \cite{shen2020interfacegan} interprets the latent semantics learned by well-trained GANs as disentangled after linear transformations such as subspace projections. We compare with InterFaceGAN because it has shown good results for precise editing control of attributes such as age, eyeglasses and expression performing semantic portrait editing. StyleFlow \cite{abdal2021styleflow} is a recent method investigating the notion of conditionally exploring the entangled GAN latent space by using conditional continuous normalizing flows. StyleFlow performs attribute-conditioned exploration and attribute-controlled editing of the latent space and enables portrait editing. We compare with StyleFlow because the normalizing flow formulation enables disentangled and fine-grained edits including expression and age. GANSpace \cite{harkonen2020ganspace} is a simple yet highly effective technique that leverages Principal Component Analysis (PCA) in the feature space or the GAN latent space to identify interpretable editing vectors by perturbing the principal edit directions. GANSpace follows a similar hypothesis as the proposed FLAME-3D in identifying disentangled edit directions in the latent space and demonstrates how the layer-wise inputs such as for the StyleGAN $\mathcal{W}+$ space can be used for editing. Hence, we compare with prior state-of-the-art editing methods: InterfaceGAN \cite{shen2020interfacegan}, StyleFlow \cite{abdal2021styleflow} and GANSpace \cite{harkonen2020ganspace}. Quantitative results are in Table(\ref{attr_table}) where the corresponding entries are obtained from previous works. Qualitative comparison of sequential edits with baselines (InterFaceGAN, StyleFlow and GANSpace) is presented in Fig.(\ref{fig:sequential})).


\paragraph{Evaluation Metrics.}
We quantitatively evaluate our method using the FID (Frechet Inception Distance) \cite{heusel2017gans}, KID (Kernel Inception Distance) \cite{binkowski2018demystifying}, ED (Euclidean Distance), and CS (Cosine Similarity) on the FFHQ dataset. FID metrics are interpreted as lower is better, ED as lower is better and CS as higher is better as indicated in Table(\ref{attr_table}). A qualitative comparison of sequential edits with baselines (InterFaceGAN, StyleFlow and GANSpace) with reference to \cite{parihar2022everything} are presented in Fig.(\ref{fig:sequential})).


\paragraph{Quantitative Results.}
We report the quantitative results of our proposed method with the respective baselines in Table(\ref{attr_table}) where the best performing values for a metric are highlighted in bold. We observe that the proposed FLAME-3D approach outperforms all baselines in FID and ED and is a very close second for the cosine similarity (CS) metric. Thus, the proposed approach is highly effective in enunciating identity preserving edits while requiring as few as 10 labeled synthetic samples.

\begin{table}
  \caption{Quantitative comparison of attribute editing. (FID = Frechet Inception Distance, ED = Euclidean Distance, CS = Cosine Similarity). Baseline results obtained from \cite{parihar2022everything}. (Best metric values are highlighted in bold).}
  \label{attr_table}
  \centering
  \begin{tabular}{c|c|c|c}
    \toprule
    Method & FID ($\downarrow$) & ED ($\downarrow$) & CS ($\uparrow$)\\
    \midrule
    InterfaceGAN \cite{shen2020interfacegan} & 43.07 & 0.61 & 0.92\\
    StyleFlow \cite{abdal2021styleflow} &47.81 & 0.71 & 0.82\\
    GANSpace \cite{harkonen2020ganspace} & 42.38 & \textbf{0.50} & \textbf{0.95}\\
    Proposed (FLAME-3D) & \textbf{39.91} & \textbf{0.50} & 0.94\\
    \bottomrule
  \end{tabular}
\end{table}

\paragraph{Qualitative Results.}
We qualitatively compare with current state-of-the-art methods in Fig.(\ref{fig:sequential}). In the figure (\ref{fig:sequential}), the first three rows correspond to results from InterfaceGAN \cite{shen2020interfacegan}, StyleFlow \cite{abdal2021styleflow} and GANSpace \cite{harkonen2020ganspace} respectively. The results correspond to sequential edits: each successive column includes the latent edit direction of the column as well as those corresponding to all previous columns. To present it more clearly, for the results corresponding to GANSpace, col 1 is the image without edits, col 2 is the face with expression modified, col 3 is expression + face aging and col 4 is expression + age + eyeglasses. We notice that the expression edit for all baseline methods seem reasonable. However, face aging and eyeglasses which correspond to difficult edits in 3D as the geometry is modified is poor in the baseline results. Specifically, we notice that the age edit for InterFaceGAN and StyleFlow change the identity of the face and the age edit for GANSpace still appears youthful with no discernible indications of aging. In comparison, the age edit corresponding to the proposed method clearly shows 3D consistent wrinkles and decreased hair quality. Moreover, the edits are 3D consistent corresponding to the angled pose. Similarly, the eyeglass edit for InterFaceGAN changes the identity, for StyleFlow there is no eyeglass added and for GANSpace, the identity, gender and hairstyle are modified with an unrealistic right ear compared to the left. Whereas the eyeglass edit for the proposed method includes 3D consistent and identity preserving edit clearly outperforming the baselines in terms of disentangled and 3D consistent latent space edit directions. 

In Fig.(\ref{fig:inv}), we show results for our experiments on inverting out-of-distribution data samples into the GAN latent space using the inversion network. We show the results using a model trained on FFHQ dataset and invert two samples from the CelebA-HQ dataset. We notice that directly inverting the image does not preserve several high frequency details of the original image as the network aims to find the closest latent that can reconstruct the input. Further, we use pivotal tuning to fine-tune the network for the given sample for only 30 steps and notice several high frequency face details to be present in the inverted image including the overall face identity and expression.

\begin{figure}
    \centering
    \includegraphics[width=0.99\linewidth]{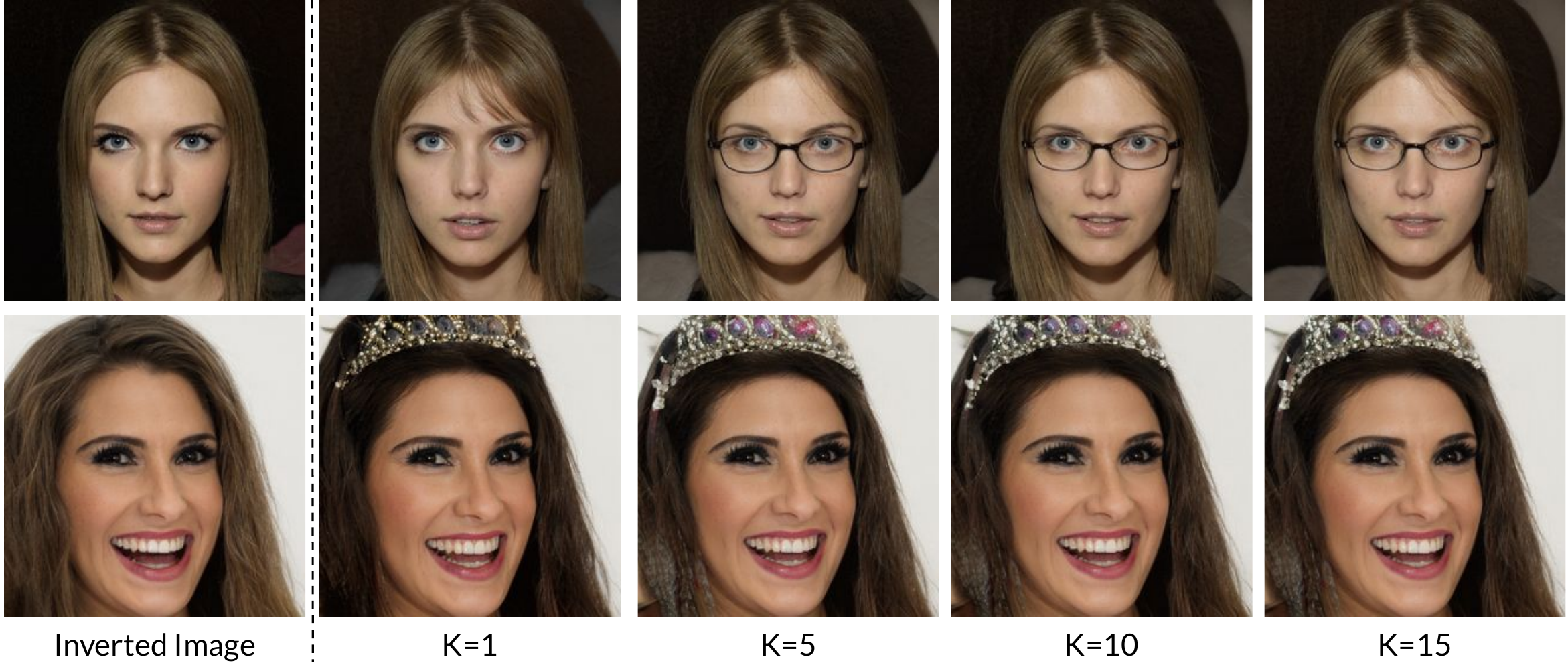}
    \caption{Ablation results for the value of K (i.e number of synthetic image pairs) required for the few-shot attribute edit direction estimation method.}
    \label{fig:ablation}
\end{figure}

\paragraph{Ablative Study.} In this work, we perform an ablation on the value of K to verify the claim that we require only upto 10 synthetic image pairs to estimate the disentangled edit directions. The qualitative results for the experiments are in Fig(\ref{fig:ablation}). Here, K refers to the number of synthetic image pairs used to compute the edit direction. We observe that for K=1 (i.e only one synthetic image pair is used), the edits are either non-existent or modifies the identity of the face. There are slight variations for K=5, the edits corresponding to using K=10 and K=15 are almost identical. Hence, we verify the claim that the proposed method requires only upto 10 synthetic image pairs to enable 3D consistent edits.

\section{Conclusion}
\label{sec:conclusion}
    In this work, we investigate an efficient and novel method of editing the latent direction of a GAN to elicit 3D and multi-view consistent edits while preserving the portrait identity. Our method draws inspiration from \cite{parihar2022everything} and uses only upto ten synthetic image pairs to compute the disentangled edit directions for attributes such as expression, eyeglasses and face aging. We demonstrate superior performance compared current state-of-the-art baselines in challenging attribute editing such as for eyeglasses and aging which includes several view dependent effects. Our method also alleviates the need for large scale paired data and semantic labels by simply copy-and-pasting edits onto an image based on the pre-trained GAN's strong prior assumption - alleviating all data bottlenecks and enabling efficient edit direction computation. The quantitative results are consistent with those reported in baseline and outperforms previous state-of-the-art works demonstrating the improved editing capabilities afforded by our method. We further verify the disentangled edits with sequential edits. Lastly, we investigate the effect of direct latent inversion and pivotal tuning inversion to preserve high frequency details in inversion and demonstrate the ability to render an image without requiring camera poses - a finding that was not possible with existing 3D-aware generative methods. In summary, to the best of our knowledge, we propose the first efficient few-shot identity preserving attribute editing method for 3D-aware deep generative models.

\section{Acknowledgements.}
\label{sec:acknowledgements}
We thank Tanmay Shah, Dmitry Lagun and Tejan Karmali for their input in helping refine the initial idea and help in debugging initial prior experiments. We also thank Tejan Karmali and Rishubh Parihar, authors of \cite{parihar2022everything} for help in verifying the claims of the proposed idea and in enabling us to include results from baseline works. Lastly, we also thank Prof. Rose Yu and Prof. Manmohan Chandraker for their support and constructive feedback that has enabled the progress of this project.

\newpage

\bibliographystyle{abbrvnat}
\bibliography{citations.bib}

\end{document}